\documentclass[10pt,twocolumn,letterpaper]{article}

\usepackage{times}
\usepackage{epsfig}
\usepackage{graphicx}
\usepackage{amsmath}
\usepackage{amssymb}
\usepackage{algorithm}
\usepackage[noend]{algpseudocode}
\usepackage{multirow}

\usepackage[pagebackref=true,breaklinks=true,colorlinks,bookmarks=false]{hyperref}

\newcommand{\tens}[1]{\boldsymbol{\mathcal{#1}}}

\begin{document}

%%%%%%%%% TITLE
\title{Tensor Yard: One-Shot Algorithm of Hardware-Friendly Tensor-Train Decomposition for Convolutional Neural Networks}

\author{Anuar Taskynov$^{2}$\\
{\tt\small taskynov.anuar@huawei.com}\\
\and
Vladimir Korviakov$^{1}$\\
{\tt\small korviakov.vladimir1@huawei.com}\\
\and
Ivan Mazurenko$^{1,2}$\\
{\tt\small mazurenko.ivan1@huawei.com}\\
\and
Yepan Xiong$^{1}$\\
{\tt\small xiongyepan@huawei.com}\\
\and
$^{1}$ {Intelligent systems and Data science Technology center, Huawei Technologies Co., Ltd, Moscow, Russia}\\
$^{2}$ {Lomonosov Moscow State University, Moscow, Russia}
}

\maketitle

%%%%%%%%% ABSTRACT
\begin{abstract}
   Nowadays Deep Learning became widely used in many economic, technical and scientific areas of human interest.
   It is clear that efficiency of solutions based on Deep Neural Networks should consider not only quality metric for 
   the target task, but also latency and constraints of target platform design should be taken into account.
   In this paper we present novel hardware-friendly Tensor-Train decomposition implementation for Convolutional Neural 
   Networks together with Tensor Yard --- one-shot training algorithm which optimizes an order of decomposition of network layers.
   These ideas allow to accelerate ResNet models on Ascend 310 NPU devices without significant loss of accuracy. For 
   example we accelerate ResNet-101 by 14.6\% with drop by 0.5 of top-1 ImageNet accuracy.
\end{abstract}

%%%%%%%%% BODY TEXT
\section{Introduction}

In recent years Deep Learning reached a significant breakthrough in many practical problems, such as Computer Vision, 
Natural Language Processing, Speech Recognition and many others.
For many years the main goal of research was to improve the quality of models, even if the model size and the latency was impractically high.
Considering the model complexity researchers typically use number of parameters or floating point operations as a complexity measure.
However, for inference of the production solutions on specific hardware like Neural Processing Unit (NPU) devices these measures are 
too abstract and are weakly connected to real latency. There exist several approaches to reduce latency of production models such as 
quantization, pruning, knowledge distillation, Neural Architecture Search (NAS) and tensor decompositions.

Multidimensional tensor in Tensor-Train (TT) format~\cite{oseledets2011tensor} is a popular tool in machine learning models.
The most known example is TensorNet~\cite{novikov_tensornet}, where parameters of linear layers in neural network are restricted to be TT tensors
of bounded ranks.
Together with automatic rank selection method which is introduced in~\cite{hawkins_autorank_tt} TensorNet 
greatly reduces the number of parameters.
In~\cite{garipov_tensornet} Tensor-Train is applied for convolutional layers and authors got comparable
results to full neural networks.
Note, that in these cases the TT format is applied to reshaped version of the parameters.
Although, due to this trick the compression of the neural network increases, it creates additional complexity
in terms of data movement. In other tensor decompositions (CP, Tucker, Tensor Ring) 
decomposed form consists of three consecutive convolutional layers (~\cite{lebedev_cp},
\cite{kim2016compression}, \cite{wang_tr}).

The other open problem of tensor decompositions is an order, in which different layers of Neural Network should be 
decomposed to reach the maximum quality of a model.

We introduce hardware-friendly Tensor-Train for convolutional layers (TTConv), which can be represented as three consecutive 
convolutional layers, where one of them is a group convolution with shared weights and two other are point-wise convolutions.
Sharing the weights between groups in the convolution help to reduce data movement operations and these weights can be stored in the device.

Secondly, we present Tensor Yard --- automatic procedure that optimizes the order of using tensor decomposition for network layers. 
This procedure consists of two steps: finding the order of decompositions and fine-tuning the model.
We replace usual convolution to weighted sum of the usual convolution and the TTConv.
After that we train the model for $M$ epochs and find the layer with the smallest corresponding weight of the usual convolution.
We replace this weighted sum to the corresponding TTConv.
This procedure continues $K$ iterations.

Proposed approach allows to free the researcher from the routine job of iterative decomposition of the model 
and accelerate it on a real hardware (we check our approach for Ascend 310 NPU device).
Due to simplicity our implementation of the TTConv can be easily reproduced.

So, our main contributions are:
\begin{itemize}
\item Novel latency-aware Tensor-Train decomposition implementation for the Convolutional Neural Networks (CNN).
\item Tensor Yard --- one-shot training procedure that optimizes the order of decomposition of the CNN layers.
\end{itemize}

%-------------------------------------------------------------------------
\section{Background and Related Art} \label{background}
Tensor decompositions is a class of methods for representation of high dimensional tensor as a sequence of 
low-cost operations to reduce the number of the tensor parameters and compress the data. 
One of the recent tensor decomposition approaches --– the Tensor-Train decomposition~\cite{oseledets2011tensor}.
The Tensor-Train decomposition generalizes the matrix low-rank format to the higher-order tensors.
A $d$-dimensional tensor $\tens{A} \in \mathbb{R}^{n_1 \times \dots \times n_d}$ is said to be represented in the Tensor-Train format (TT-format for 
short) if each of its elements $\tens{A}_{i_1, \dots, i_d}$
equals to the following product of $2$ vectors and $d-2$ matrices:
\begin{equation}
\begin{aligned}
  \label{TT-repr}
  \tens{A}_{i_1, i_2, \dots, i_d} = G_1[i_1] \times G_2[i_2] \times \dots \times G_d[i_d],
\end{aligned}
\end{equation}
The 3-dimensional arrays $G_\mu \in \mathbb{R}^{r_{\mu-1} \times n_\mu \times r_\mu}$, $\mu = 2,\dots,d-1$ for any value of $i_\mu=1,\dots,n_\mu$
return a $r_{\mu-1}\times r_{\mu}$ matrix $G_\mu[i_\mu]$. The first and the last arrays are 2-dimensional and for any value of $i_1=1,\dots,n_1$ and
$i_d = 1, \dots, n_d$ return a vector $G_1[i_1] \in \mathbb{R}^{1\times r_1}$, $G_d[i_d] \in \mathbb{R}^{r_{d-1} \times 1}$.
These arrays define the decomposition and are called the "TT-cores". The vector $\boldsymbol{r} = (1, r_1, \dots, r_{d-1}, 1)$ of the slices of the 
TT-cores is called the "TT-rank" of the decomposition.
In order to represent a large vector in the TT-format, it's reshaped to multidimensional tensor by factorizing 
dimension of the vector and~\eqref{TT-repr} is applied.

For the matrix $\boldsymbol{W}$ the TT-format is given by:
\begin{equation}
\begin{aligned}
\label{tt-matr-format}
\tens{W}_{i_1, i_2, \dots, i_d; j_1, j_2, \dots, j_d} = \\
= G_1[i_1, j_1] \times G_2[i_2, j_2] \times \dots \times G_d[i_d, j_d],
\end{aligned}
\end{equation}
where $\tens{W}$ is a reshaped form of $\boldsymbol{W}$.
This representation is also known as the "TT matrix".

The TT-format for a tensor with $n^d$ elements and the TT-rank $r$ requires only $(d-2)nr^2 + 2nr$ parameters (elements of the TT-cores) to store,
which make the TT-format very efficient when the TT-rank $r$ is small. To find $G_1, G_2,\dots, G_d$ there exists the TT-SVD algorithm, 
which is a recursive generalization of SVD for so-called tensor unfoldings.

In~\cite{novikov_tensornet} the TT format has been applied to neural networks.
Authors use the TT format to reduce number of the parameters
in fully-connected layers by reshaping weights to multidimensional tensor and applying~\eqref{tt-matr-format}.

Tensor-train decomposition for convolutions (TTConv) has been proposed by~\cite{garipov_tensornet}.
To use the TT matrix format to $l\times l$ convolutional layer authors do the following steps: 
\begin{itemize}
\item Factorize input and output channels: $C = \prod_{i=1}^{d} C_i$, $S = \prod_{i=1}^d S_i$
\item Define a $(d+1)$-dimensional tensor $K$, where dimension length is $C_k S_k$, if $k \in \{1, \dots, d\}$ and $l^2$, 
if $k = 0$;
\item Consider that $c$ and $s$ are indices of corresponding reshaped indices $\{c_i \}_{i=1}^d$, $\{s_i \}_{i=1}^d$;
\item The TT-matrix representation of convolutional kernel tensor:\\
\begin{equation}
\begin{gathered}
\mathcal{K}[i, j, c, s] = \tens{K}[i, c_1 \dots, c_d; j, s_1, \dots, s_d] = \\
G_0[i, j] \times G_1[c_1, s_1] \times \dots \times G_d[c_d, s_d]
\end{gathered}
\end{equation}
\item The TTConv operation can be defined as:
\begin{equation}
\begin{gathered}
\boldsymbol{Y}[h, w, s_1, \dots, s_d] = \\
\sum_{i=1,j=1}^{l,l} \sum_{c_1, \dots, c_d} \boldsymbol{X}[h', w', c_1, \dots, c_d] \cdot \\
\cdot \big(G_0[i, j] \times G_1[c_1, s_1] \times \dots \times G_d[c_d, s_d]\big).
\end{gathered}
\end{equation}
\end{itemize}

As mentioned above, there exist methods (~\cite{lebedev_cp}, \cite{kim2016compression}, \cite{wang_tr}) which transform the convolutional kernel into the 3-dimensional tensor (the kernel size dimensions
are combined) and obtain 3 consecutive convolutions by applying decomposition:
\begin{itemize}
   \item $1 \times 1$ convolution from input channels to $R_1$;
   \item $l \times l$ special form of convolution from $R_1$ to $R_2$;
   \item $1 \times 1$ convolution from $R_2$ to output channels.
\end{itemize}
By special form of convolution we mean usual convolution (\cite{kim2016compression}), depth-wise convolution (\cite{lebedev_cp}) or group convolution (\cite{wang_tr}).

\section{Problem}

In this section we discuss the design of the NPU hardware on the example of Ascend 310 NPU publicly available on the market~\cite{huawei2021ascend310} and constraints on the CNN operations and tensor decompositions design.
After that, we investigate an existing implementations of Tensor-Train decomposition.
Finally we pay attention to the problem of an optimal layers decomposition order, which solution, as we show, is important to reach high accuracy of the model.
\begin{figure*}
\begin{center}
\includegraphics[width=0.8\linewidth]{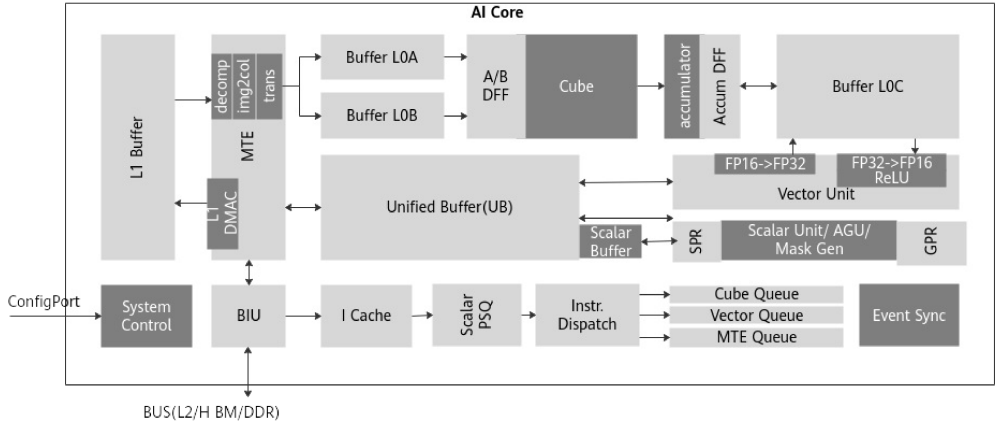}
\end{center}
\caption{Da Vinci AI Core architecture~\cite{huawei2020davinci}.}
\label{fig:aicore}
\end{figure*}

\subsection{NPU design and constraints}

   In recent years there appeared a lot of AI acceleration hardware and these devices set certain limitations for the models to be deployed.
Examples of such domain-specific devices are: Google Cloud TPU~\cite{tpu2019edge, 10.1145/3154484}, NVidia Jetson~\cite{nvidiajetson}, Huawei Ascend~\cite{huawei2021ascend310}, Intel Movidius Myriad~\cite{intelmovidiusmyriad} and many others.
These devices are typically good at parallelizable tasks of tensor and matrix multiplications and additions as well as other operations commonly used in Neural Networks, such as activation functions and other element-wise operations.\\
\\
In our paper we consider optimization for Ascend 310 NPU based on Da Vinci architecture~\cite{huawei2020davinci}, 
however, our solution is wide enough to be applicable for any AI accelerators that support standard operations like 2D convolution with groups.\\
\\
Overall design of Da Vinci AI Core is shown in figure~\ref{fig:aicore}.
AI Core is the main part of Ascend 310 NPU, and it executes tensor and vector operations.
Three main compute units of AI Core are: Cube Unit, which performs matrix multiplications, including fully-connected layers and convolutions;
Vector Unit which executes vector operations like element-wise sum of tensors, Batch Normalization~\cite{batchnorm} and activation functions;
Scalar Unit, which is responsible for scalar operations and controls program flow and addressing.\\
\\
Cube Unit performs multiplication of two 16x16 fp16 matrices or 16x32 and 32x16 int8 matrices at a time, and this is one of the most important constraint imposed by the design of the Cube Unit.
Matrices of larger size are multiplied by parts.
If a size of multiplied matrices is less than specified they will be padded by zeros.
It is acceptable, but the highest Cube Unit utilization is reached for the matrices with a size divisible by 16 or 32 depending on computation precision.\\
\\
Vector Unit is responsible for the vector computations. It provides less computational power than Cube Unit, but the capabilities of computations are more flexible.\\
\\
To process and store data there are several storage units in AI Core, including L1 Buffer (general internal storage of a large size), L0 Buffers (storage of input and output data for the Cube Unit), Unified Buffer, etc.
Memory Transfer Engine (MTE) manages read/write operations between different buffers and performs operations like padding, transposition and img2col.\\
\\
Control Units of AI Core (e.g. System Control and Queues) provide instruction control for the computation process.\\
\\
What conclusions can be made to design efficient Neural Networks architectures and tensor decomposition schemes for NPU?
First of all, for all shapes of tensors (including weights and activations) that are processed by the Cube Unit divisibility by 16 is preferable.
Secondly, the matrix operations are more preferable than the vector operations and operations with data.
When possible it is better to avoid operations like element-wise product, sum and permutations.
Third, the tensor decomposition should not significantly increase the chain of operations.
Every operation in the computational graph requires input and output data to be transferred which increases total latency and may cancel positive effect of the decomposition.

\begin{figure*}
\begin{center}
\includegraphics[width=1.0\linewidth]{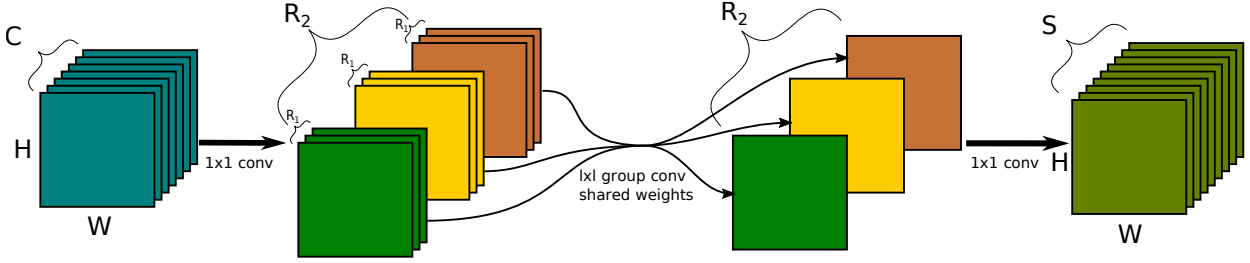}
\end{center}
\caption{Graphical representation of the proposed TTConv operation.}
\label{fig:newttconv}
\end{figure*}

\subsection{Current implementation of Tensor-Train decomposition for convolution}

Current application of the TT format to convolution, which is described in section~\ref{background} requires data permutation operation, which is not optimal for the NPU devices.
In the perfect scenario decomposition should be implemented using the most efficient operations for the NPU devices, which can be reduced to the matrix multiplications, such as 
fully-connected layers and convolutions.

\subsection{Optimal decomposition order}

The other important problem is the selection of the order for the sequential decomposition of the Neural Network’s layers, which is the NP-hard 
problem.
To our knowledge, all of the existing approaches don’t take into consideration which layers are good to be decomposed and the order 
of decomposition of these layers.
Typically, the decomposed models are either trained from scratch or approximation methods like the TT-SVD or the VBMF~\cite{kim2016compression} 
are used.
In the MUSCO approach ~\cite{gusak_musco} the authors propose an iterative algorithm of the gradual redundancy 
reduction with the automatic rank selection.

\section{Approach}

\subsection{Hardware-friendly Tensor-Train decomposition}
There exist two main sources of the model’s slowdown in real hardware: computation (tensor, matrix, vector and scalar operations) and data processing (input and output data transfer, memory operations such as tensor permutations).
Many of the existing tensor decomposition approaches require transposition or, more generally, permutation operations.
These operations require expensive memory operations and should be avoided when possible.
\\
We propose a new TTConv operation, where we apply the Tensor-Train decomposition for the 3-dimensional tensor, which is a reshape of the usual convolutional tensor, where kernel size 
dimensions are combined.
So, the TT decomposition is applied to $l \times l$ convolution from $S$ channels to $C$ channels as follows:
\begin{equation}
\begin{gathered}
\mathcal{K}_{s, c, i, j} = G_1[i,j] \times G_2[c, 1] \times G_3[s, 1] = \\
= \sum_{r_1, r_2=1}^{R_1, R_2} G_1[1, i, j, r_1] \cdot G_2[r_1, c, 1, r_2] \cdot G_3[r_2, s, 1, 1],
\end{gathered}
\end{equation}
where  $G_1 \in \mathbb{R}^{1\times l \times l \times R_1}$, $G_2 \in \mathbb{R}^{R_1 \times C \times 1 \times R_2}$, 
$G_3 \in \mathbb{R}^{R_2, S, 1, 1}$ --- TT Matrix cores. After omitting the redundant indices we can obtain:
\begin{equation}
\begin{gathered}
   \mathcal{K}_{s, c, i, j} = \sum_{r_1, r_2=1}^{R_1, R_2} G_1[i, j, r_1] \cdot G_2[r_1, c, r_2] \cdot G_3[r_2, s].
\end{gathered}
\end{equation}
In this case, the convolutional layer is represented as:
\begin{equation}
\begin{gathered}
\boldsymbol{Y}_{h, w, s} = \\
= \small{\sum_{c=1}^C \sum_{i, j=1}^{l, l} \sum_{r_1,r_2=1}^{R_1, R_2} \boldsymbol{X}_{h', w', c}
\cdot \big(G_1[i, j, r_1] \cdot G_2[r_1, c, r_2] \cdot G_3[r_2, s]\big)}.
\end{gathered}
\end{equation}
After rearranging the multipliers TTConv can be represented as the 3 convolutional layers, where the second convolution is a group convolution with shared kernel weights (figure~\ref{fig:newttconv}):
\begin{itemize}
\item $1 \times 1$ convolution from $C$ channels to $R_1 R_2$ channels;
\item $l \times l$ group convolution with groups=$R_2$  from $R_1 R_2$ channels to $R_2$, where the convolutional kernel weight is the same for all groups;
\item $1 \times 1$ convolution from $R_2$ channels to $S$ channels.  
\end{itemize}

This decomposition of the convolutional weight is applicable only if $l > 1$ and in the case of $1 \times 1$ convolution we use the low-rank decomposition as for the matrices:
\begin{equation}
\begin{aligned}
\boldsymbol{Y}_{h, w, s} = \sum_{c=1}^C \sum_{r=1}^R \boldsymbol{X}_{h, w, c} \cdot G_1[c, r] \cdot G_2[r, s].
\end{aligned}
\end{equation}

As it was mentioned before, the Cube Unit performs the multiplication with 16x16 matrices and divisibility of the ranks by 16 is preferable, so we use $R_2=16$ and $R_1 = \frac{C}{4  \cdot R2}$ for $l \times l$ convolutions and $R=16$ for $1x1$ convolutions.
We apply the tensor decompositions to all convolutional layers, with the number of channels $\ge 128$.

Our implementation of Tensor-Train Convolution has the following computational complexity:
\begin{equation}
\begin{aligned}
\mathcal{C} = O\Big(HW \cdot (C R_1 R_2 + R_1 R_2 l^2 + R_2 S) \Big)
\end{aligned}
\end{equation}
And the memory complexity is:
\begin{equation}
\begin{aligned}
\mathcal{C} = O\Big(C R_1 R_2 + R_1 l^2 + R_2 S \Big)
\end{aligned}
\end{equation}
where $H$ and $W$ - height and width of data; $C_{in}$ - number of input channels; $S$ - number of output channels; $l$ - size of convolutional kernel; $R_1$ and $R_2$ - Tensor-Train ranks.

Proposed implementation has the following advantages:
\begin{itemize}
\item It uses only the standard group and the pointwise convolutional operations supported and highly optimized by all modern Deep Learning frameworks and the hardware.
\item It can be easily incorporated into the modern convolutional architectures like the ResNet.
\item It can be further optimized using the lower-level capabilities of the AI acceleration API.
\end{itemize}

\subsection{Tensor Yard One-Shot Training Procedure}

\begin{figure}[t]
\begin{center}
\includegraphics[width=1.0\linewidth]{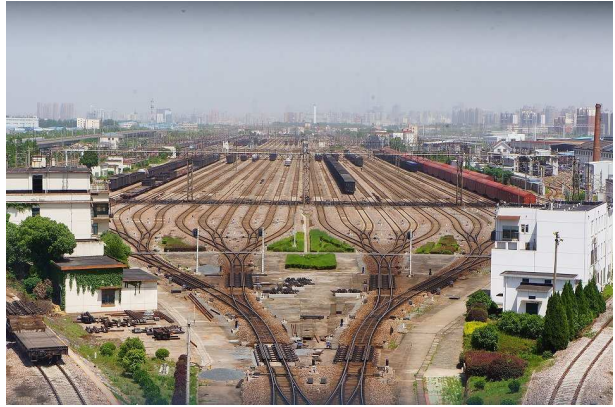}
\end{center}
   \caption{Train classification yard is an analogy for our one-shot algorithm~\cite{yard2018}.}
\label{fig:yard}
\end{figure}

To address the problem of the automatic selection of the layers decomposition order we propose the one-shot training algorithm.
We call it the "Tensor Yard" by an analogy with a train classification yard, which is used to separate and reorder the train cars (figure ~\ref{fig:yard}).
This algorithm is inspired by the differentiable Neural Architecture Search (NAS) algorithms~\cite{liu2018darts,chen2019progressive}. 
This family of the NAS algorithms find architectures by the joint optimization of the overperemeterized network parameters and trainable coefficients, assigned to different paths in the architecture.
The Tensor Yard training algorithm is started from replacing of all usual convolutions to the $\alpha$-weighted sums of usual convolution and Tensor-Train decomposition of the convolution, the TTConv, weight $\alpha$ is trainable parameter and $\alpha \in [0..1]$:
\begin{equation}
op(x,\alpha) \gets \alpha_i \cdot Conv(x)+(1-\alpha_i) \cdot TTConv(x) \label{eq:1}
\end{equation}
Consider the case when $\alpha$ corresponds to the usual convolution in the weighted sum.
Thus, each layer has trainable weight $\alpha$ which indicates the importance of the usual convolution layer.
If $\alpha$ is close to zero, then using the TTConv is more suitable.
In the training process we sort the layers by coefficient $\alpha$ and find the lowest $\alpha$.
It means that in this iteration, switching usual convolution to the TTConv is more appropriate.
Pseudo-code of our algorithm is given in Algorithm~\ref{tensoryard}.
Generally, our approach is not limited to the Tensor-Train decomposition only, but we leave the use of other decomposition methods for the future work.

\begin{algorithm}
   \caption{Tensor Yard algorithm}\label{tensoryard}
   \begin{algorithmic}[1]
   \Procedure{TensorYard}{}
   \State {$L \gets $ Number of layers to decompose}
   \State {$M \gets $ Number of epochs in iteration}
   \State {$K \gets $ Number of iterations}
   \For{$i \gets 1$ to $L$} 
      \State $\alpha_{i} \gets 0.5$
      \State $op_{i}(x,\alpha_{i}) \gets \alpha_i \cdot Conv_{i}(x)+(1-\alpha_i) \cdot TTConv_{i}(x)$
   \EndFor

   \For{$j \gets 1$ to $K$}
      \State {Train model for $M$ epochs}
      \State $l \gets \underset{i}{\mathrm{argmin}}(\alpha_{i})$
      \If{$\alpha_l < 0.5$}
         \State $op_{l}(x,\alpha_{l}) \gets TTConv_{l}(x)$
      \EndIf
   \EndFor

   \For{each remaining $op_{i}(x,\alpha_{i})$}
      \State $op_{i}(x,\alpha_{i}) \gets Conv_{i}(x)$
   \EndFor

   \EndProcedure
   \end{algorithmic}
\end{algorithm}

Proposed approach releases the researcher from the routine job of selection of the decomposition order and solves this problem automatically.

%-------------------------------------------------------------------------
\section{Experiments}

For all of our experiments we use the PyTorch framework and the Automatic Mixed Precision (O2) training and $5$ epochs of the warmup~\cite{goyal_warmup}. Initial learning rate is equal to $0.1*BatchSize/256$. 
As architectures we use the classical ResNets~\cite{he_resnet}. During the Tensor Yard learning rate does not change, after that we use:
\begin{itemize}
   \item cosine learning rate schedule for the CIFAR datasets;
   \item step schedule (reducing learning rate $10$ times after each $30$ epochs) for the ImageNet dataset.
\end{itemize}
After the Tensor Yard our compressed models are trained for $90$ epochs.

\subsection{Ablation Study}

\begin{figure*}
\begin{center}
\includegraphics[width=1.0\linewidth]{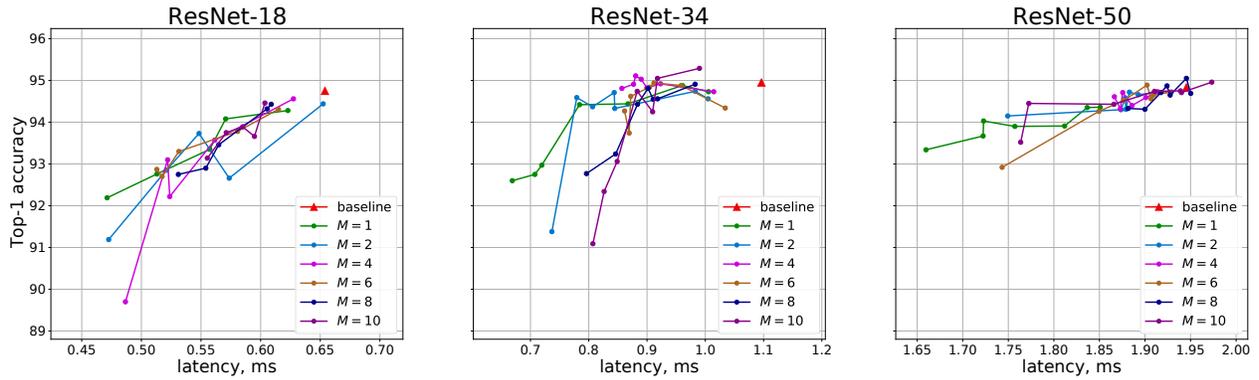}
\end{center}
\caption{Impact of parameter $M$ to training result on CIFAR-10 dataset}
\label{fig:cifar10_ablation}
\end{figure*}

\begin{figure*}
\begin{center}
\includegraphics[width=1.0\linewidth]{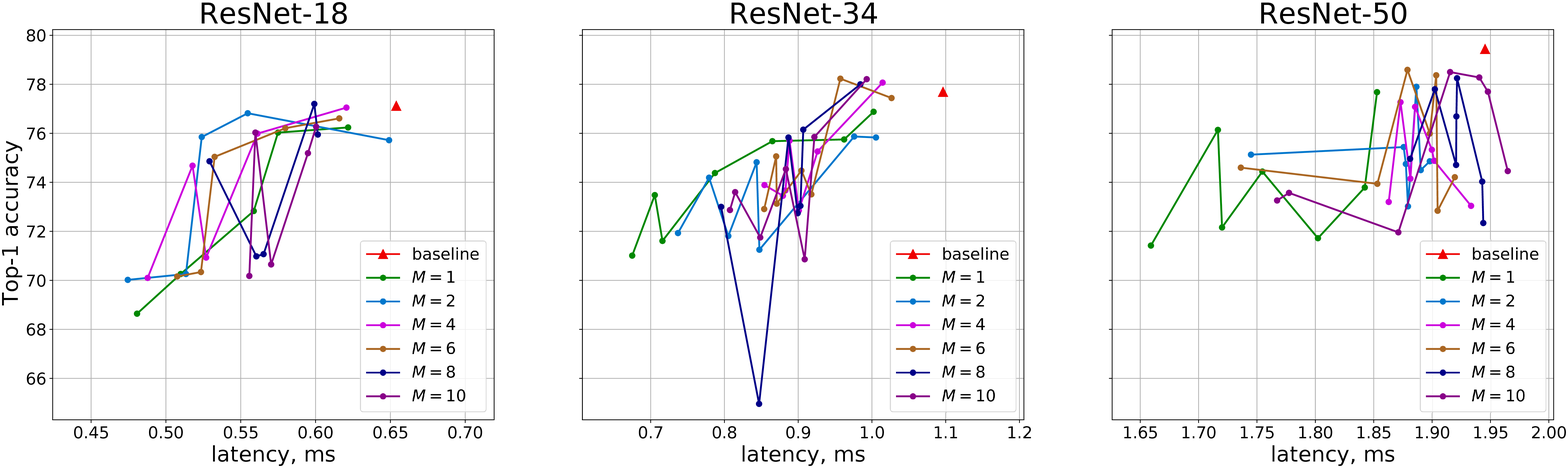}
\end{center}
\caption{Impact of parameter $M$ to training result on CIFAR-100 dataset}
\label{fig:cifar100_ablation}
\end{figure*}

To study an impact of the parameter $M$ (the number of epoch per iteration of the Tensor Yard algorithm) we train ResNet-18, 34 and 50 models on CIFAR-10 and 
CIFAR-100 datasets with $M \in [1, 2, 4, 6, 8, 10]$.
The results are shown in figures~\ref{fig:cifar10_ablation} and ~\ref{fig:cifar100_ablation} for CIFAR-10 and CIFAR-100 correspondingly.
It can be seen that good accuracy/acceleration trade-off can be achieved on the small values of $M$: 1 to 6 epochs per 
iteration are typically enough for the training.
Moreover, for the high values of $M$ the algorithm tends to select non-decomposed layers, thus, not reach the highest possible acceleration.
The latency of all resulting architectures are measured on the Ascend 310 NPU with the batch size 32 and reduced to a single image (i.e. end-to-end time of batch processing is divided by the size of batch).

\begin{table}[b]
\centering
\begin{tabular}{|c|c|c|c|c|c|c|c|c|c|}
\hline
\multirow{3}{*}{Model} & \multicolumn{6}{c|}{Latency (ms)} & \multirow{3}{*}{\#Param.} & \multirow{3}{*}{GFLOPs} & \multirow{3}{*}{Top-1 Acc. (\%)} \\
\cline{2-7}
& \multicolumn{2}{c|}{Batch 8} & \multicolumn{2}{c|}{Batch 16} & \multicolumn{2}{c|}{Batch 32} & & &\\
\cline{2-7}
& time & $\Delta$ (\%) & time & $\Delta$ (\%) & time & $\Delta$ (\%) & & &  \\
\hline\hline
ResNet-18 & 0.655 & - & 0.629 & - & 0.654 & - & 11M & 1.8 & 69.76 \\
\hline

ResNet-18-SFP-0.7 & 0.589 & 5.45 & 0.545 & 13.3 & 0.54 & 17.4 & 8.4M & 1.32 & 67.10 \\
\hline
ResNet-18-VBMF & 0.555 & 15.19  & 0.521 & 17.13 & \textbf{0.504} & \textbf{22.88} & 3.9M & 1.2 & 65.628 \\
\hline
ResNet-18-TY (ours) & 0.623 & 4.89 & 0.585 & 7 & 0.592 & 9.48 & 7M & 1.59 & \textbf{69.29} \\
\hline\hline
ResNet-34 & 1.071 & - & 1.035 & - & 1.096 & - & 21.8M & 3.6 & 73.3 \\
\hline
ResNet-34-SFP-0.7 & 0.921 & 14 & 0.854 & 17.48 & 0.862 & 21.3 & 15.5M & 2.6 & 71.83 \\
\hline
ResNet-34-VBMF & 0.886 & 17.26 & \textbf{0.819} & \textbf{20.82} & 0.81 & 26.02 & 7.6M & 2.3 & 70.94 \\
\hline
ResNet-34-TY (ours) & 0.916 & 14.47 & 0.873 & 15.65 & 0.910 & 16.97 & 17.3M & 2.5 & \textbf{73.15} \\
\hline\hline
ResNet-50 & 1.7 & - & 1.695 & - & 1.945 & - & 25.5M & 4.08 & 76.15 \\
\hline
ResNet-50-SFP-0.7 & 1.56 & 8.24 & 1.61 & 5.01 & 1.75 & 10.02 & 16.9M & 2.6 & 74.61 \\
\hline
ResNet-50-VBMF & 1.555 & 8.5 & 1.527 & 9.89 & 1.743 & 10.39 & 17.78M & 3.4 & 75.13 \\
\hline
ResNet-50-TY (ours) & 1.5 & 11.76 & \textbf{1.492} & \textbf{11.98} & 1.724 & 11.36 & 16.9M & 3.2 & \textbf{75.18} \\
\hline\hline
ResNet-101 & 2.676 & - & 2.607 & - & 2.917 & - & 44.5M & 7.8 & 78 \\
\hline
ResNet-101-SFP-0.7 & 2.30 & 14.05 & 2.35 & 9.86 & 2.591 & 11.18 & 28.2M & 4.8 & 77.51 \\
\hline
ResNet-101-VBMF & 2.284 & 14.66 & \textbf{2.163} & \textbf{17.03} & 2.382 & 18.33 & 29.12M & \textbf{5.5} & \textbf{77.734} \\
\hline
ResNet-101-TY (ours) & 2.342 & 12.48 & 2.224 & 14.69 & 2.497 & 14.39 & 30.9M & 6.13 & 77.53 \\
\hline
\end{tabular}
\caption{Experiments on ImageNet dataset. "TY" denotes TensorYard (our approach)}
\label{table:imagenet_result}
\end{table}

\subsection{Results on ImageNet}

The results of the experiments on the ImageNet dataset are shown in the table~\ref{table:imagenet_result}. We compare our approach with other methods including the Soft Filter Pruning (SFP)~\cite{he2018soft} and the VBMF with the Tucker decomposition~\cite{kim2016compression}.
Existing approaches accelerate ResNet-18 and 34 well, but the accuracy drop is much higher (up to 2.36 and 4.14) while our method reduces the accuracy by less than 0.5. For ResNet-50 our method has better both the speed and the accuracy.
For ResNet-101 our approach is slightly worse, but still comparable with the VBMF.
Thus, our method provides better accuracy/latency trade-off than the other approaches making it valuable for the real applications and it was the main goal of our research.
The latency of all resulting architectures are measured on the Ascend 310 NPU with batch size 8, 16 and 32 and reduced to a single image.

\section{Conclusions}

In this paper we present our implementation of the Tensor-Train decomposition for the convolutional layers.
This implementation is friendly to the NPU design and can bring the acceleration on real hardware.
Proposed simple scheme of the decomposition can be easily incorporated to existing architectures as the replacement of the conventional convolutional layers.
The other contribution is the one-shot training algorithm inspired by the differentiable NAS approaches.
This algorithm addresses the problem of the optimal order of the CNN layers decomposition and provides an automatic 
solution of this problem and saves the researcher's time and efforts.\\
Our experience shows that the application of the tensor decompositions (and, specifically, the Tensor-Train) for acceleration of
Neural Networks on real hardware is a complicated task and requires the theoretical analysis of the decomposition algorithm,
but also constraints of the target device should be taken into account.
\\
As a direction for the future work and improvement of our approach we consider an automation of the optimal 
hardware-friendly rank selection (currently we heuristically predefine all ranks of the decomposition) and integration with 
other acceleration approaches like pruning.

%-------------------------------------------------------------------------
\newpage

%{\small
%\bibliographystyle{ieee_fullname}
%\bibliography{egbib}
%}

\bibliographystyle{ieee_fullname.bst}
\bibliography{egbib}

\end{document}